\documentclass{tlp}
\usepackage{latexsym}
\usepackage{times}

\title{\bf Propositional Defeasible Logic has \\ Linear Complexity}

\author{M.J.\ Maher \\
Department of Mathematical and Computer Sciences \\
Loyola University Chicago \\
{\em mjm@cs.luc.edu} \\
\ \\
School of Computing \& Information Technology \\
Griffith University \\
}

\date{}

\begin{document}

\maketitle

\newcommand{\ignore}[1]{}
\newcommand{\comment}[1]{}
\newcommand{\finish}[1]{}

\newtheorem{thm}{Theorem}
\newtheorem{lemma}{Lemma}
\newtheorem{definition}{Definition}
\newtheorem{defn}{Definition}
\newtheorem{corollary}{Corollary}
\newtheorem{proposition}{Proposition}
\newtheorem{propn}{Proposition}
\newtheorem{auxexample}{Example}
\newenvironment{example}{\begin{auxexample} }{\ $\Box$\end{auxexample}}
\ignore{
\newenvironment{example}{\begin{auxexample}\em }{\ $\Box$\end{auxexample}}
\newenvironment{proof}{\begin{quotation} \noindent \em
	   {\bf Proof:\ }}{\(\Box\) \end{quotation}}

\newcommand{\qed}{
    \setlength{\unitlength}{6pt}
    \begin{picture}(1,1)
    \thicklines
    \multiput(0,0)(0.0,0.01){100}{\line(1,0){1}}
    \end{picture}
}

}

\newcommand{\ps}[1]{+\sigma #1}
\newcommand{\ms}[1]{-\sigma #1}
\newcommand{\pd}[1]{+\partial #1}
\newcommand{\md}[1]{-\partial #1}
\newcommand{\PD}[1]{+\Delta #1}
\newcommand{\MD}[1]{-\Delta #1}

\newcommand{\p}{{\em ~~Poss~~}}
\newcommand{\np}[1]{{\em ~~NP(#1)~~}}

\newcommand{\non}{\!\sim\!\!}

\newcommand{\QADL}{{\it Deimos}}
\newcommand{\ACDL}{Delores}

\begin{abstract}
Defeasible logic is a rule-based nonmonotonic logic,
with both strict and defeasible rules, and a priority relation on rules.
We show that inference in the propositional form of the logic
can be performed in linear time.
This contrasts markedly with most other propositional nonmonotonic logics,
in which inference is intractable.

\end{abstract}

\section{Introduction} \label{intro}

\newcommand{\PP}{$\Pi^p_2$}

Most work in non-monotonic reasoning has focussed on languages
for which propositional inference is not tractable.
Sceptical default reasoning is \PP-hard,
even for very simple classes of default rules,
as is sceptical autoepistemic reasoning
and propositional circumscription.
The complexity of sceptical inference from
logic programs with negation-as-failure
varies according to the semantics of negation.
For both the stable model semantics and the Clark completion,
sceptical inference is co-NP-hard.
See \cite{Gottlob92,CS} for more details.

Although such languages are very expressive,
and this expressiveness has been exploited in 
answer-set programming \cite{Niemela}, for example,
they have not led to any practical applications in
non-monotonic reasoning.

There has also been work on more tractable languages.
Extensive work on inheritance networks with exceptions
has led to polynomial time algorithms \cite{Stein}
and applications \cite{Morgenstern98}.
Defeasible logic \cite{Nute87,Nute94}
is a generalization \cite{Billington90} of 
inheritance networks with exceptions
under the directly sceptical semantics \cite{Horty}.
Defeasible logic replaces the implicit specificity relation
by an explicit, programmable priority relation;
generalizes containment statements between concepts and their complements
to rules over literals;
and adds the notion of an explicit defeater.
Recent work has proposed other languages \cite{Grosof97,DK}
which are essentially subsets of defeasible logic \cite{AMB}.

We have already established that full first-order defeasible logic
has a recursively enumerable inference problem \cite{MG99}.
In this paper we establish that inference in propositional defeasible logic
has linear complexity.
Considering the several expressive features of defeasible logic,
and its low complexity,
it appears to be a suitable basis for applications of non-monotonic
reasoning that involve very large rule sets.

In the next section we introduce the proof theory of
defeasible logic, which defines the logic.
Then we present a transition system 
(for a subset of defeasible logic)
that progressively simplifies a defeasible theory while
accumulating conclusions that can be inferred from the theory.
We show that the meaning of the theory is preserved,
and when no more transitions are applicable
all consequences have been accumulated.

The following section presents an algorithm that can be viewed
as performing a particular sequence of transitions.
By an appropriate choice of data structures, we show that
the set of all conclusions can be computed in time
linear in the size of the theory.
Finally, we use the transformations of \cite{ABM98,ABGM99}
to map an arbitrary defeasible theory to the subset of
defeasible logic to which the algorithm applies.
These transformations require only linear time,
and blowup in the size of the theory is linear.
Thus the result is a linear time computation of
the conclusions of the defeasible theory.
This algorithm has been the basis of an implementation
of defeasible logic \cite{MRABM}.

\section{Defeasible Logic} \label{sec:dl}

\subsection{An Informal Presentation}
\label{sec:informal}
We begin by presenting the basic ingredients of defeasible logic. A
defeasible theory (a knowledge base in defeasible logic, or a
defeasible logic program)
consists of five different kinds of knowledge: facts, strict rules,
defeasible rules, defeaters, and a superiority relation.

\smallskip\noindent
{\em Facts} are indisputable statements, for example, ``Tweety is an
emu''.  Written formally, this would be expressed as $emu(tweety)$.

\smallskip\noindent
{\em Strict rules} are rules in the classical sense: whenever the
premises are indisputable (e.g.\ facts) then so is the conclusion. An
example of a strict rule is ``Emus are birds''. Written formally:
\[
emu(X) \rightarrow bird(X).
\]

\smallskip\noindent
{\em Defeasible rules} are rules that can be defeated by contrary
evidence. An example of such a rule is ``Birds typically
fly''; written formally: 
\[
bird(X) \Rightarrow flies(X).
\]
The idea is that if we know that something is a bird, then we may
conclude that it flies, {\em unless there is other, not inferior, 
evidence suggesting that it may not fly}.

\smallskip\noindent 
{\em Defeaters} are rules that cannot be used to
draw any conclusions. Their only use is to prevent some conclusions.
In other words, they are used to defeat some defeasible rules by
producing evidence to the contrary. An example is ``If an animal is
heavy then it might not be able to fly''. Formally:
\[
heavy(X) \leadsto \neg flies(X).
\]
The main point is that the information that an animal is heavy is not
sufficient evidence to conclude that it does not fly. It is only
evidence that the animal {\em may} not be able to fly. In other words,
we don't wish to conclude $\neg flies(X)$ if $heavy(X)$, we simply
want to prevent a conclusion $flies(X)$
unless there is more evidence that overrides the heaviness of $X$.

\smallskip\noindent
The {\em superiority relation} among rules is used to define
priorities among rules, that is, where one rule may override the
conclusion of another rule.  For example, given the defeasible rules
\[
\begin{array}{lrl}
r: & bird(X) & \Rightarrow flies(X) \\

r':  & brokenWing(X) & \Rightarrow \neg flies(X) \\
\end{array}
\]
which contradict one another, no conclusive decision can be made about
whether a bird with broken wings can fly. But if we introduce a
superiority relation $>$ with $r'>r$, 
with the intended meaning that $r'$ is strictly stronger than $r$,
then we can indeed conclude that
the bird cannot fly.

There are several relevant points to be made about
the superiority relation in defeasible logic.
Notice that a cycle in the
superiority relation is counter-intuitive.  In the above example, it
makes no sense to have both $r>r'$ and $r'>r$.  Consequently, 
defeasible logic requires that the superiority relation is acyclic.

A second point worth noting is that, in defeasible logic, priorities
are {\em local} in the following sense: Two rules are considered to be
competing with one another exactly when they have complementary
heads. Thus, since the superiority relation is used to resolve conflicts
among competing rules, it is only used to compare rules
with complementary heads; the information $r>r'$ for rules $r,r'$
without complementary heads may be part of the superiority relation,
but has no effect on the proof theory. 

Finally, sets of rules with competing heads work as {\em teams} against
each other.
For example,
consider the following rules about mammals
\[
\begin{array}{lrl}
r_1: & monotreme(X) & \Rightarrow mammal(X) \\

r_2: & hasFur(X)    & \Rightarrow mammal(X) \\

r_3: & laysEggs(X)  & \Rightarrow \neg mammal(X) \\

r_4: & hasBill(X)   & \Rightarrow \neg mammal(X) \\

\end{array}
\]
with the following superiority relation:
$r_1 > r_3$
and
$r_2 > r_4$.
For a platypus, which satisfies all antecedents,
no rule is superior to all competing rules.
However, rules $r_1$ and $r_2$ together are superior to rules $r_3$ and $r_4$.
Hence defeasible logic concludes that a platypus is a mammal.

Defeasible logic has some similarities with default logic \cite{Reiter80}:
both are non-monotonic logics that distinguish statements which are unarguable
(strict rules/ facts)
from statements that have lesser force (defeasible rules/defaults).
However, there are substantial syntactic and semantic differences.

First,
default logic has no notion of priority among rules,
nor of a defeater that cannot be used to make inferences.
On the other hand, default logic permits any logical expression as a fact
or in a default rule,
whereas defeasible logic allows only rules as statements.
Nevertheless, even if we restrict attention to propositional default theories
$(W, D)$ where $W$ is a conjunction of literals and
$D$ consists of Horn default rules
(so that the syntax is weaker than defeasible logic),
sceptical inference is co-NP-hard \cite{CS}
whereas, as will be shown in this paper,
defeasible logic has linear complexity.

As noted by Nute \cite{Nute94},
a central difference between the two logics is in the application of rules.
In default logic the applicability of a default rule is
independent of any other default rule so that,
given default rules
${~:~ a} \over {a}$
and
${~:~ \neg a} \over {\neg a}$,
either default may be applied.
On the other hand, in defeasible logic with rules
$\{ \Rightarrow a, \Rightarrow \neg a\}$
neither rule may be applied, since each rule ``interferes'' with
-- or defeats -- the other.

\subsection{Formal Definition}
\label{sec:formal}

In this paper we restrict attention to essentially propositional
defeasible logic. Rules with free variables can be interpreted as rule
schemas, that is, as the set of all variable-free instances;
in such cases we assume that the Herbrand universe is finite
and we regard variable-free atoms as propositions.
For compactness of presentation, we use rule schemas in examples.
However, in this paper we assume that a theory is presented
in propositional form, and not with rule schemas.

We assume that
the reader is familiar with the notation and basic notions of
propositional logic.
If $q$ is a literal, $\non q$ denotes the
complementary literal (if $q$ is a positive literal $p$ then $\non q$
is $\neg p$; and if $q$ is $\neg p$, then $\non q$ is $p$).

A {\em rule} $r:A(r)\hookrightarrow C(r)$ consists of its unique
\emph{label} $r$, its {\em antecedent} (or {\em body}) $A(r)$
($A(r)$ may be omitted if it is the empty set)
which is a finite set of literals, an arrow
$\hookrightarrow$ (which is a placeholder for concrete arrows
to be introduced in a moment),
and its  {\em consequent} (or {\em head}) $C(r)$ which is a literal.
In writing rules
we omit set notation for antecedents and sometimes we omit the
label when it is not relevant for the context.  There are three kinds
of rules, each represented by a different arrow.  Strict rules use
$\rightarrow$, defeasible rules use $\Rightarrow$, and defeaters use
$\leadsto$.

Given a set $R$ of rules, we denote the
set of all strict rules in $R$ by $R_s$,
the set of strict and defeasible rules in $R$ by $R_{sd}$,
the set of defeasible rules in $R$ by $R_d$,
the set of defeasible rules and defeaters in $R$ by $R_{dd}$,
and the set of defeaters in $R$ by $R_{dft}$. 
$R[q]$ denotes the set of rules in $R$ with consequent $q$.

A {\em superiority relation on $R$} is a relation $>$ on $R$.  When
$r_1>r_2$, then $r_1$ is called {\em superior} to $r_2$, and $r_2$
{\em inferior} to $r_1$.  Intuitively, $r_1>r_2$ expresses that $r_1$
overrules $r_2$, should both rules be applicable. We assume
$>$ to be acyclic (that is, the transitive closure of $>$ is
irreflexive).

A {\em defeasible theory} $D$ is a triple $(F,R,>)$ where $F$ is a
finite set of literals (called {\em facts}), $R$ a finite set of
rules, and $>$ an acyclic superiority relation on $R$.

\newcommand{\bird}{{\sl bird}}
\begin{example} \label{bird}
We will use the following defeasible theory to demonstrate several aspects
of defeasible logic.
We assume there are only the constants $ethel$ and $tweety$
in the language.
Let
$D_\bird = (F_\bird, R_\bird, >_\bird)$
where:
$F_\bird$ is the set of facts
\[
\begin{array}{lrrl}
& & & emu(ethel). \\
& & & bird(tweety). \\
\end{array}
\]
$R_\bird$ is represented by the set of rule schemas
\[
\begin{array}{lrrl}
r_1: & emu(X) & \rightarrow & bird(X). \\
r_2: & bird(X) & \Rightarrow & flies(X). \\
r_3: & heavy(X) & \leadsto & \neg flies(X). \\
r_4: & brokenWing(X) & \Rightarrow & \neg flies(X). \\
r_5: & & \Rightarrow & heavy(ethel). \\
\end{array}
\]
and the superiority relation $>_\bird$
contains only $r_4 >_\bird r_2$.

The five rule schemas
give rise to
nine propositional rules
by instantiating each variable to $ethel$ and $tweety$ respectively.
Those rules, which make up $R_\bird$ are
\[
\begin{array}{lrrl}
r_{1,e}: & emu(ethel) & \rightarrow & bird(ethel). \\
r_{1,t}: & emu(tweety) & \rightarrow & bird(tweety). \\
r_{2,e}: & bird(ethel) & \Rightarrow & flies(ethel). \\
r_{2,t}: & bird(tweety) & \Rightarrow & flies(tweety). \\
r_{3,e}: & heavy(ethel) & \leadsto & \neg flies(ethel). \\
r_{3,t}: & heavy(tweety) & \leadsto & \neg flies(tweety). \\
r_{4,e}: & brokenWing(ethel) & \Rightarrow & \neg flies(ethel). \\
r_{4,t}: & brokenWing(tweety) & \Rightarrow & \neg flies(tweety). \\
r_{5}: & & \Rightarrow & heavy(ethel). \\
\end{array}
\]
The rules have been re-labelled purely to simplify later reference\footnote{
Without the re-labelling we would have different rules with the same label.
This is not a problem, formally, but might be confusing.
}.
As a result, the superiority relation becomes
$\{
r_{4,e}  >_\bird r_{2,e},
r_{4,e}  >_\bird r_{2,t},
r_{4,t}  >_\bird r_{2,e},
r_{4,t}  >_\bird r_{2,t}
\}
$.
As noted in the previous subsection, the two statements
$r_{4,e}  >_\bird r_{2,t}$
and
$r_{4,t}  >_\bird r_{2,e}$
have no effect,
since they do not involve rules with conflicting heads.

In this defeasible theory,
$R_s = \{ r_{1,e}, r_{1,t} \}$
and $R_d[\neg flies(tweety)] = \{ r_{3,t}, r_{4,t} \}$.
\end{example}

\subsection{Proof Theory}
\label{sec:prooftheory}
Defeasible logic is defined by the proof theory presented
in this subsection.
There are also characterizations of defeasible logic
in other frameworks \cite{MG99,GM00,Mah00},
but they are not needed here.

A {\em conclusion} 
of $D$ is a tagged literal that may be proved by $D$,
and can have one of the following four forms:

\begin{itemize}
  
\item[ $+\Delta q$] which is intended to mean that $q$ is definitely
  provable in $D$.
  
\item[$-\Delta q$] which is intended to mean that we have proved
  that $q$ is not definitely provable in $D$.
  
\item[$+\partial q$] which is intended to mean that $q$ is
  defeasibly provable in $D$.
  
\item[$-\partial q$] which is intended to mean that we have proved
  that $q$ is not defeasibly provable in $D$.
\end{itemize}

Conclusions are used only at the meta-level and do not occur in
defeasible theories.
Notice the distinction between $-$, which is used only to express
unprovability, and $\neg$, which expresses classical negation.
For example,
$\MD{\neg flies(tweety)}$ means
that it has been proved that the negated proposition $\neg flies(tweety)$
cannot be proved definitely in the defeasible theory.

If we are able to prove $q$
definitely, then $q$ is also defeasibly provable. This is a direct
consequence of the formal definition below.
Similarly, if $-\partial q$ is concluded
then we must also conclude $-\Delta q$.

Provability is defined below. It is based on the concept of a {\em
  derivation} (or {\em proof}) in $D=(F,R,>)$. A derivation is a finite
sequence $P=(P(1),\ldots P(n))$ of tagged literals constructed by
inference rules.
There are four inference rules
(corresponding to the four kinds of conclusion)
that specify how a derivation may be extended.
($P(1..i)$ denotes the initial part of the
sequence $P$ of length $i$):

\begin{tabbing}
90123456\=7890\=1234\=5678\=9012\=3456\=\kill

\>$+\Delta$: We may append $P(i+1) = +\Delta q$ if either \\
\>\>\> $q\in F$ or \\
\>\>\> $\exists r\in R_s[q] \ \forall a\in A(r): +\Delta a \in
P(1..i)$ 
\end{tabbing}

This means, to prove $+\Delta q$ we need to establish a proof for $q$
using facts and strict rules only. This is a deduction in the
classical sense -- no proofs for the negation of $q$ need to be
considered (in contrast to defeasible provability below, where
opposing chains of reasoning must also be taken into account).
From $D_\bird$ in Example \ref{bird}
we can infer $\PD{emu(ethel)}$ (and $\PD{bird(tweety)}$) immediately,
in a proof of length 1.
Using $r_{1,e}$ and the second clause of the inference rule,
we can infer $\PD{bird(ethel)}$ in a proof of length 2.

\begin{tabbing}
90123456\=7890\=1234\=5678\=9012\=3456\=\kill
\> $-\Delta$: We may append $P(i+1)=-\Delta q$ if \\
\>\>\> $q\not\in F$ and \\
\>\>\> $\forall r\in R_{s}[q] \ \exists a\in A(r): -\Delta a \in P(1..i)$
\end{tabbing}

To prove $-\Delta q$, that is, that $q$ is not definitely provable, $q$
must not be a fact. In addition, we need to establish that every
strict rule with head $q$ is {\em known to be} inapplicable. Thus for
every such rule $r$ there must be at least one element $a$ of the antecedent
for which we have established that $a$ is not definitely provable
($-\Delta a$).
From $D_\bird$, we can infer \linebreak
$\MD{heavy(tweety)}$ and $\MD{\neg flies(tweety)}$
immediately (among many others), since in these cases $R_{s}[q]$ is empty.

Note that this definition of nonprovability does not
involve loop detection. Thus if $D$ consists of the single rule
$p\rightarrow p$, we can see that $p$ cannot be proven, but defeasible
logic is unable to prove $- \Delta p$.

Defeasible provability requires consideration of chains of reasoning
for the complementary literal, and possible resolution using the superiority
relation.
Thus the inference rules for defeasible provability
are more complicated than those for definite provability.

\begin{tabbing}
90123456\=7890\=1234\=5678\=9012\=3456\=\kill
\>$+\partial$: \> We may append $P(i+1)=+\partial q$ if either \\
\>\>(1) $+\Delta q \in P(1..i)$ or \\
\>\>(2) \> (2.1) $\exists r\in R_{sd}[q] \forall a \in
A(r): +\partial a\in P(1..i)$ and \\
\>\>\>(2.2) $-\Delta\non q \in P(1..i)$ and \\
\>\>\>(2.3) $\forall s \in R[\non q]$ either \\
\>\>\>\>(2.3.1) $\exists a\in A(s): -\partial a\in P(1..i)$ or \\
\>\>\>\>(2.3.2) $\exists t\in R_{sd}[q]$ such that \\
\>\>\>\>\> $\forall a\in A(t): +\partial a\in P(1..i)$ and $t>s$ 
\end{tabbing}

Let us illustrate this definition. To show that $q$ is provable
defeasibly we have two choices: (1) We show that $q$ is already
definitely provable; or (2) we need to argue using the defeasible part
of $D$ as well. In particular, we require that there must be a strict
or defeasible rule with head $q$ which can be applied (2.1). But now
we need to consider possible ``attacks'', that is, reasoning
chains in support of $\non q$. To be more specific: to prove $q$
defeasibly we must show that $\non q$ is not definitely provable
(2.2). Also (2.3) we must consider the set of all rules which are not
known to be inapplicable and which have head $\non q$ (note that here
we consider defeaters, too, whereas they could not be used to support
the conclusion $q$; this is in line with the motivation of defeaters
given in subsection 2.1). Essentially each such rule $s$ attacks the
conclusion $q$. For $q$ to be provable, each such rule $s$ must be
counterattacked by a rule $t$ with head $q$ with the following
properties: (i) $t$ must be applicable at this point, and (ii) $t$
must be stronger than $s$. Thus each attack on the conclusion $q$ must
be counterattacked by a stronger rule.

From $D_\bird$ in Example \ref{bird}
we can infer $\pd{bird(ethel)}$ in a proof of length 3, using (1).
Other applications of this inference rule first require
application of the inference rule $-\partial$.
This inference rule completes
the definition of the proof theory of defeasible logic.
It is a strong negation of the inference rule $+\partial$ \cite{ABGM00}.

\begin{tabbing}
90123456\=7890\=1234\=5678\=9012\=3456\=\kill

\>$-\partial$: \> We may append $P(i+1)=-\partial q$ if  \\
\>\>(1) $-\Delta q \in P(1..i)$ and \\
\>\>(2) \> (2.1) $\forall r\in R_{sd}[q] \ \exists a \in
A(r): -\partial a\in P(1..i)$ or \\
\>\>\>(2.2) $+\Delta\non q \in P(1..i)$ or \\
\>\>\>(2.3) $\exists s \in R[\non q]$ such that \\
\>\>\>\>(2.3.1) $\forall a\in A(s): +\partial a\in P(1..i)$ and \\
\>\>\>\>(2.3.2) $\forall t\in R_{sd}[q]$ either \\
\>\>\>\>\> $\exists a\in A(t): -\partial a\in P(1..i)$ or $t\not> s$ 
\end{tabbing}

To prove that $q$ is not defeasibly provable, we must first establish
that it is not definitely provable. Then we must establish that it
cannot be proven using the defeasible part of the theory. There are
three possibilities to achieve this: either we have established that
none of the (strict and defeasible) rules with head $q$ can be applied
(2.1); or $\non q$ is definitely provable (2.2); or there must be an
applicable rule $s$ with head $\non q$ such that no possibly applicable rule
$t$ with head $q$ is superior to $s$ (2.3).

From $D_\bird$ in Example \ref{bird}
we can infer $\md{brokenWing(ethel)}$ with a proof of length 2
using (1) and (2.1),
since $R[q]$ is empty.
Employing this conclusion, we can
then infer $\md{\neg flies(ethel)}$, again using (2.1),
since the only rule $r$ is $r_{4,e}$.
Using $r_5$,
we can infer $\pd{heavy(ethel)}$ by (2) of the inference rule $\pd{}$;
(2.3) holds since $R[\non q]$ is empty.
Using (2.3) of the inference rule $\md{}$,
we can infer $\md{flies(ethel)}$,
where the role of $s$ is taken by $r_{3,e}$.

Furthermore,
we can infer $\md{brokenWing(tweety)}$ and $\md{heavy(tweety)}$
using (2.1) of inference rule $\md{}$,
since in both cases $R_{sd}[q]$ is empty.
Employing those conclusions,
we can then infer $\md{\neg flies(tweety)}$, again using (2.1),
since the only rules for $\neg flies(tweety)$ involve antecedents
already established to be unprovable defeasibly.
Now, using this conclusion, we can use (2), specifically (2.1), of
the inference rule $\pd{}$ to infer $\pd{flies(tweety)}$.

The elements of a derivation 
$P$ in $D$
are called {\em lines} of the derivation.
We say that a tagged literal $L$ is {\em provable} in $D=(F,R,>)$,
denoted $D\vdash L$, iff there is a derivation in $D$ such that $L$ is
a line of $P$.  
Equivalently, we say that $L$ is a conclusion (or {\em consequence})
of $D$.

Conclusions are the basis of our notion of equivalent defeasible theories.
We say $D_1$ and $D_2$ are {\em conclusion equivalent}
(written $D_1 \equiv D_2$) iff
$D_1$ and $D_2$ have identical sets of consequences, that is,
$D_1 \vdash L$ iff $D_2 \vdash L$.

An important property of defeasible logic is that it is {\em coherent},
that is, there is no defeasible theory $D$
and literal $p$ such that
$D \vdash \pd{p}$ and $D \vdash \md{p}$, or
$D \vdash \PD{p}$ and $D \vdash \MD{p}$ \cite{Billington93}.
Put simply, this property says that we cannot establish that
a literal is simultaneously provable and unprovable.

Notice that strict rules are
used in two different ways. When we try to establish {\em definite
provability}, then strict rules are used as in classical logic: if
their bodies are proved definitely then their head is proved definitely,
regardless of any reasoning chains
with the opposite conclusion. But strict rules can also be used to
show {\em defeasible provability}, given that some other literals are
known to be defeasibly provable. In this case, strict rules are used
exactly like defeasible rules. For example, a strict rule may
have its body proved defeasibly, yet it may not fire because
there is a  rule with the opposite conclusion 
that is not weaker. 
Furthermore, strict rules are not automatically
superior to defeasible rules.
In this paper,
we will choose to duplicate strict rules as defeasible rules,
and require definite reasoning to use the strict rules (as always),
while defeasible reasoning may use only defeasible rules.
Because of the above facts, this duplication and separation of rules
does not modify the consequences.
When there is a duplicate defeasible rule for every strict rule
in a defeasible theory, we say that the theory has
{\em duplicated strict rules}.

In this paper we will focus on the subset of defeasible logic that involves
no superiority statements and no defeaters,
which we call {\em basic defeasible logic}.
For basic defeasible logic
the inference rules are simplified \cite{ABGM99}.
We simplify them further by introducing two auxiliary tags for literals
($\ps$ and $\ms$) and corresponding inference rules.

\begin{tabbing}
90123456\=7890\=1234\=5678\=9012\=3456\=\kill
\>$+\sigma$: We may append $P(i+1) = +\sigma q$ if \\
\>\>\> $\exists r\in R_{sd}[q] \ \forall a\in A(r): +\partial a \in
P(1..i)$
\end{tabbing}

\begin{tabbing}
90123456\=7890\=1234\=5678\=9012\=3456\=\kill
\> $-\sigma$: We may append $P(i+1)=-\sigma q$ if \\
\>\>\> $\forall r\in R_{sd}[q] \ \exists a\in A(r): -\partial a \in P(1..i)$
\end{tabbing}

These tags ($\ps{}, \ms{}$) represent the ability/inability to find
a tentative reason for the literals.
Essentially, we can prove $\ps{q}$ when there is an argument for $q$,
but we must consider all counter-arguments (for $\:\non q$)
before we can prove $\pd{q}$.
Similarly, we can prove $\ms{q}$ if there is not even
a tentative reason for $q$,
but there are other ways to conclude $\md{q}$.
Tagged literals involving $\sigma$, $\partial$ or $\Delta$
are called {\em extended conclusions}.

With the addition of these tags and inference rules,
the inference rules for $\partial$ are reduced to:

\begin{tabbing}
90123456\=7890\=1234\=5678\=9012\=3456\=\kill
\>$+\partial$: We may append $P(i+1) = \pd{q}$ if either \\
\>\>\> $\PD{q} \in P(1..i)$ or \\
\>\>\> $\{ \ps{q}, \MD{\non q}, \ms{\non q} \}  \subseteq P(1..i)$ 
\end{tabbing}

\begin{tabbing}
90123456\=7890\=1234\=5678\=9012\=3456\=\kill
\>$-\partial$: We may append $P(i+1) = \md{q}$ if \\
\>\>\> $\MD{q} \in P(1..i)$ and \\
\>\>\> $\{ \ms{q}, \PD{\non q}, \ps{\non q} \}  \cap P(1..i) \neq \emptyset$ 
\end{tabbing}

This modified proof system
reduces the amount of ``work'' done in any one proof step
in comparison with the original inference rules,
and is the logical basis for the algorithm.
However, the algorithm is also based on repeated simplification
of the defeasible theory,
which we describe as a transition system in the next section.

\section{The Transition System} \label{trans}

\newcommand{\arr}{\Longrightarrow}
\newcommand{\notarr}{\;\not\!\!\Longrightarrow}

The algorithm is defined in terms of a transition system
on states.
A {\em state} is a pair $(D, C)$ of a defeasible theory $D$
and a set of extended conclusions $C$.
As the algorithm proceeds, the theory $D$ is simplified
and conclusions of the theory are accumulated in $C$.
The transitions for the positive conclusions
are based on forward chaining.
The negative conclusions are derived by a dual process.

\begin{defn}
There is a transition $(D_i, C_i) \arr (D_{i+1}, C_{i+1})$
only in the following cases:

\begin{enumerate}
\item  \label{PD}
$(D_i, C_i) \arr (D_i, C_i \cup \{ \PD{q}, \pd{q}, \ps{q} \})$ \\
if there is a fact $q$ in $D_i$ or
a strict rule in $D_i$ with head $q$ and empty body
 
\item  \label{ps}
$(D_i, C_i) \arr (D_i, C_i \cup \{ \ps{q} \})$ \\
if there is a defeasible rule in $D_i$ with head $q$ and empty body
 
\item  \label{MD}
$(D_i, C_i) \arr (D_i, C_i \cup \{ \MD{q} \})$ \\
if there is no strict rule in $D_i$ with head $q$
and no fact $q$ in $D_i$

\item  \label{ms}
$(D_i, C_i) \arr (D_i, C_i \cup \{ \ms{q} \})$ \\
if there is no rule in $D_i$ with head $q$

\item  \label{inferpd}
$(D_i, C_i) \arr (D_i, C_i \cup \{ \pd{q} \})$ \\
if $\PD{q} \in C_i$ or
if
$\{  \MD{\non q}, \ms{\non q}, \ps{q} \} \subseteq C_i$

\item  \label{infermd}
$(D_i, C_i) \arr (D_i, C_i \cup \{ \md{q} \})$ \\
if $\MD{q} \in C_i$ and either
$\PD{\non q} \in C_i$ or
$\ps{\non q} \in C_i$ or 
$\ms{q} \in C_i$.

\item  \label{delPD}
$(D_i, C_i) \arr ((D_i \backslash \{r\}) \cup \{r'\}, C_i)$ \\
if $r$ is a strict rule in $D_i$ with body containing $q$,
$r'$ is $r$  with $q$ deleted,
and $\PD{q} \in C_i$

\item  \label{delpd}
$(D_i, C_i) \arr ((D_i \backslash \{r\}) \cup \{r'\}, C_i)$ \\
if $r$ is a defeasible rule in $D_i$ with body containing $q$,
$r'$ is $r$ with $q$ deleted,
and $\pd{q} \in C_i$

\item  \label{delMD}
$(D_i, C_i) \arr ((D_i \backslash \{r\}), C_i)$ \\
if $r$ is a strict rule in $D_i$ with body containing $q$,
and $\MD{q} \in C_i$

\item  \label{delmd}
$(D_i, C_i) \arr ((D_i \backslash \{r\}), C_i)$ \\
if $r$ is a defeasible rule in $D_i$ with body containing $q$,
and $\md{q} \in C_i$

\end{enumerate}

We write $(D_i, C_i) \notarr$
if there is no transition $(D_i, C_i) \arr (D_{i+1}, C_{i+1})$
except those where $(D_i, C_i) = (D_{i+1}, C_{i+1})$.

If
$(D, \emptyset) \arr \cdots \arr (D', C') \notarr$
and $C$ is the subset of $C'$ involving only the tags 
$\PD{}, \MD{}, \pd{}$, and $\md{}$,
then we say that $D$ {\em derives} $C$.
\end{defn}

We demonstrate the action of the transition system on
$D_\bird$ in Example \ref{bird}.
Transition \ref{PD} applies to each fact
and transition \ref{MD} applies to all literals
except the facts and $bird(ethel)$.
Similarly, transition \ref{ps} applies to $heavy(ethel)$
and transition \ref{ms} applies to several literals, 
including all $brokenWing$ literals.
These transitions do not modify $D_i$;
they only accumulate conclusions in $C_i$.
However, as a result of this accumulation,
transition \ref{delPD} deletes $emu(ethel)$ from $r_{1,e}$,
which then allows transition \ref{PD} to add $\pd{bird(ethel)}$
to $C_i$, and
transition \ref{delMD} deletes $r_{1,t}$.
Furthermore,
transition \ref{inferpd} applies for $emu(ethel)$, $bird(tweety)$, and
$bird(ethel)$ by the first condition,
and then transition \ref{delpd} applies to modify
$r_{2,e}$ and $r_{2,t}$;
similarly, transition \ref{infermd} applies to all $brokenWing$ literals,
which then allows
$r_{4,e}$ and $r_{4,t}$ to be deleted by transition \ref{delmd}.
Further transitions are possible.

It is quite clear that, by ignoring defeaters and superiority statements,
the transition system is incorrect for theories of full defeasible logic.
The transition system is also incomplete, in general, for theories of
basic defeasible logic.
For example, from the rules
\[
\begin{array}{rl}
a & \rightarrow b \\
  & \Rightarrow a \\
\end{array}
\]
the transition system cannot conclude $\ps{b}$.
But under the assumption that strict rules are duplicated as defeasible rules,
the transition system is sound and complete, as we now show.

\begin{thm}
Let $D$ be a basic defeasible theory with duplicated strict rules.
Suppose
$(D, \emptyset) \arr \cdots \arr (D', C') \notarr$.
If $c \in C'$ then $D \vdash c$.

\begin{proof}
Consider a transition $(D_i, C_i) \arr (D_{i+1}, C_{i+1})$.
We prove, by induction on $i$, that
\begin{itemize}
\item
If $c \in C_{i+1} \backslash C_i$ then $D_i \vdash c$
\item
$D_i \equiv D_{i+1}$
\end{itemize}
for every conclusion $c$.

The first part is straightforward, since the first six transitions
are essentially instances of the inference rules of defeasible logic.
The remainder of the transitions do not modify $C_i$.

For the second part only the last four transitions are relevant.
We need to show that, for each transition and every proof in $D_i$
there is a corresponding proof in $D_{i+1}$, and vice versa.

{\em For transition \ref{delPD}}: \\
If a proof $p$ in $D_i$ applies the inference rule $\PD{}$ 
using $r$, then the inference rule is still applicable using $r'$
and $p$ is also a proof in $D_{i+1}$.
Similarly,
if a proof $p$ in $D_i$ applies the inference rule $\MD{}$ 
involving $r$, then $p$ establishes $\MD{x}$,
for some $x$ in the body of $r$.
$x$ cannot be $q$, since $D_i \vdash \PD{q}$.
Thus $p$ is also a proof in $D_{i+1}$.

If a proof $p$ in $D_{i+1}$ applies the inference rule $\PD{}$ 
using $r'$, let $p'$ be a proof in $D_i$ of $\PD{q}$.
Then the concatenation of $p'$ and $p$ is a proof in $D_i$.
If a proof $p$ in $D_{i+1}$ applies the inference rule $\MD{}$ 
involving $r'$, then $p$ is also a proof in $D_i$.

{\em For transition \ref{delpd}}: \\
Let $s$ be the head of $r$.
If $p$ is a proof in $D_i$ containing $\ps{s}$
then $p$ is also a proof in $D_{i+1}$.
If $p$ is a proof in $D_{i+1}$ containing $\ps{s}$,
let $p'$ be a proof in $D_i$ of $\pd{s}$.
Then the concatenation of $p'$ and $p$ is a proof in $D_i$.

If $p$ is a proof in $D_i$ containing $\ms{s}$,
then there is some literal $t$ in the body of $r$
such that $D_i \vdash \md{t}$.
$t$ cannot be $q$ since, by assumption, $D_i \vdash \pd{q}$,
and defeasible logic is coherent.
Thus $p$ is also a proof in $D_{i+1}$.
If $p$ is a proof in $D_{i+1}$ containing $\ms{s}$,
then there is some literal $t$ in the body of $r'$
such that $D_{i+1} \vdash \md{t}$.
We must also have $D_{i} \vdash \md{t}$,
since the proof in $D_{i+1}$ cannot use $r'$,
which would introduce circularity.
Thus $p$ is also a proof in $D_i$.

{\em For transition \ref{delMD}}: \\
If a proof $p$ in $D_{i}$ applies the inference rule $\PD{}$ 
it does not use $r$ since $D_i \vdash \MD{q}$.
Thus $p$ is also a proof in $D_{i+1}$.
If a proof $p$ in $D_i$ applies the inference rule $\MD{}$ 
then the inference rule also applies in $D_{i+1}$.
Thus $p$ is also a proof in $D_{i+1}$.

If a proof $p$ in $D_{i+1}$ applies the inference rule $\PD{}$ 
then it is also a proof in $D_{i}$.
If a proof $p$ in $D_{i+1}$ applies the inference rule $\MD{}$ 
let $p'$ be a proof in $D_i$ of $\MD{q}$.
Then the concatenation of $p'$ and $p$ is a proof in $D_i$.

{\em For transition \ref{delmd}}: \\
Let $s$ be the head of $r$.
If $p$ is a proof in $D_i$ containing $\ms{s}$
then $p$ is also a proof in $D_{i+1}$.
If $p$ is a proof in $D_{i+1}$ containing $\ms{s}$,
let $p'$ be a proof in $D_i$ of $\md{q}$.
Then the concatenation of $p'$ and $p$ is a proof in $D_i$.

If $p$ is a proof in $D_{i}$ containing $\ps{s}$,
then the proof of $\ps{s}$ cannot use $r$, since $D_i \vdash \md{q}$.
Thus $p$ is also a proof in $D_{i+1}$.
If $p$ is a proof in $D_{i+1}$ containing $\ps{s}$,
then $p$ is also a proof in $D_{i+1}$.
\end{proof}
\end{thm}

It follows from the above proof that the transformed
defeasible theories are equivalent.
\begin{corollary}
Let $D$ be a basic defeasible theory with duplicated strict rules.
Suppose
$(D, \emptyset) \arr \cdots \arr (D', C')$.
Then $D \equiv D'$.
\end{corollary}

The transition system is complete:
it generates all the conclusions that are inferred by
defeasible logic.
\begin{thm}
Let $D$ be a basic defeasible theory with duplicated strict rules.
Suppose
$(D, \emptyset) \arr \cdots \arr (D', C') \notarr$.
For every conclusion $c$,
if
$D' \vdash c$ then $c \in C'$.

\begin{proof}
Suppose, to obtain a contradiction, that the theorem does not hold,
and let $c$ be a conclusion of $D'$ that is not in $C'$
and has a minimal length proof in $D'$ among such conclusions.

If $c$ is $\MD{q}$ then 
every strict rule with head $q$ has a body literal $b$
such that $D' \vdash \MD{b}$.
By the assumption,
and since the proof for each $\MD{b}$ must be shorter,
$\MD{b} \in C'$.
Since no more transitions are possible,
and considering transition \ref{delMD},
there is no strict rule with head $q$.
But then, a transition \ref{MD} that adds $\MD{q}$ to $C'$,
is possible, contradicting the assumption.

If $c$ is $\PD{q}$ then 
there is a strict rule with head $q$ and $D' \vdash \PD{b}$
for each body literal $b$.  
By the assumption,
and since the proof for $\PD{b}$ must be shorter,
$\PD{b} \in C'$.
Since no more transitions are possible,
and considering transition \ref{delPD},
the body of the rule must be empty.
But then, considering transition \ref{PD}, $\PD{q} \in C'$,
contradicting the assumption.

Similarly, if $c$ is $\pd{q}$ then there must be a defeasible rule for
$q$ with an empty body.
Furthermore, $D' \vdash \MD{\non q}$,
and every defeasible rule for $\non q$ contains a body literal $b$
such that $D' \vdash \md{b}$.
By the assumption,
$\md{b} \in C'$, for each $b$, and $\MD{\non q} \in C'$.
Since no more transitions are possible,
considering transition \ref{delmd}
there must be no defeasible rules for $\non q$.
Thus $\ms{\non q} \in C'$.
Also $\ps{q} \in C'$, since the rule body is empty,
and thus the transition \ref{inferpd} applies,
leading to a contradiction.

If $c$ is $\md{q}$ then
$D' \vdash \MD{q}$, and either
every defeasible rule for $q$ contains a body literal $b$
such that $D' \vdash \md{b}$
or
$D' \vdash \PD{\non q}$
or
and there is a defeasible rule for $\non q$ such that 
for all its body literals $b'$, 
$D' \vdash \pd{b'}$.
By the assumption,
and since any of these proofs must be shorter than the proof of $c$,
$\MD{q} \in C'$ and either
every defeasible rule for $q$ contains a body literal $b$
such that $\md{b} \in C'$
or
$\PD{\non q} \in C'$
or
and there is a defeasible rule for $\non q$ such that 
for all its body literals $b'$, 
$\pd{b'} \in C'$.
But then, since no transitions are possible,
and considering transition \ref{delpd} and \ref{delmd},
either there are no defeasible rules for $q$
or
$\PD{\non q} \in C'$
or there is a defeasible rule for $\non q$ with an empty body.
Considering transitions \ref{ps} and \ref{ms}, either
$\ms{q} \in C'$
or
$\PD{\non q} \in C'$
or
$\ps{\non q} \in C'$.
Consequently, because of transition \ref{infermd},
$\md{q} \in C'$,
contradicting the original assumption.
\end{proof}
\end{thm}

Combining the previous two theorems, we can conclude that
the transition system
computes exactly the conclusions of 
the initial defeasible logic program.

\begin{thm}
Let $D$ be a basic defeasible theory with duplicated strict rules.
Suppose $D$ derives $C$.
Then, for every conclusion $c$,

$D \vdash c$ iff $c \in C$
\end{thm}

\section{The Linear Algorithm} \label{acdl}

\newcommand{\To}{\Rightarrow}
\newcommand{\es}{\mathit{elim\_sup}}

\newcommand{\Basic}{$\mbox{\sl Basic}$}
\newcommand{\DupStrict}{$\mbox{\sl DupStrict}$}
\newcommand{\CheckInference}{$\mbox{\sl CheckInference}$}

The algorithm consists of two main parts:
(1)
a series of transformations to produce an equivalent
basic defeasible theory with duplicated strict rules, and
(2)
the application of transitions from the transition system
in a computationally efficient order.
These parts will be discussed in separate subsections.

\subsection{Transformations}

The input to the algorithm is an arbitrary defeasible theory,
but the transition system is proved correct only for
basic defeasible theories with duplicated strict rules.
To bridge this gap we must transform the input defeasible theory
into the appropriate form.

In \cite{ABGM99} three transformations are  used to convert any
defeasible theory into an equivalent basic defeasible theory.
The first places the defeasible theory in a normal form,
the second eliminates defeaters,
and the third reduces the superiority relation to the empty relation.
Let \Basic{} denote this sequence of transformations.
We will use \Basic{} in the inference algorithm.

It is beyond the scope of this paper to present a definition of 
\Basic{};
for that, the reader is referred to \cite{ABGM99}.
However,
to give the flavor of the transformations
we present one component transformation, $\es$,
which empties the superiority relation.
It introduces two new propositions for each rule,
two rules for each superiority statement,
and up to four rules replacing each original rule.
It is defined as follows.

\begin{defn}
Let $D$ be a regular defeasible theory with rules $R$.
Let $\Sigma$ be the language of $D$. 
\[
  \begin{array}{lrl}
    \es(R)=R_{s} & \cup ~ \{&\!\neg inf^{+}(r_1)\To inf^{+}(r_2),\\
       		 &			&\!\neg inf^{-}(r_1)\To inf^{-}(r_2) ~|~ r_1>r_2\} \\
       		 & \cup ~ \{&A(r)\To \neg inf^{+}(r),\\
         	 &			&\neg inf^{+}(r)\To p,\\
         	 &			&A(r)\To \neg inf^{-}(r),\\
         	 &			&\neg inf^{-}(r)\To p ~|~ r\in R_d[p]\}\\
       		 & \cup ~ \{&A(r)\To \neg inf^{-}(r),\\
         	 &			&\neg inf^{-}(r)\leadsto p ~|~ r\in R_{dft}[p]
       \}
  \end{array}
\]
For each $r$, $inf^{+}(r)$ and $inf^{-}(r)$ are new atoms not in
$\Sigma$. Furthermore all new atoms are distinct.
\end{defn}

Intuitively, the two introduced propositions, $inf^{+}(r)$ and $inf^{-}(r)$,
represent that $r$ is inferior to a rule in $R_{sd}$ (respectively $R$)
that has a defeasibly provable body.
If $D \vdash \ps{inf^{+}(r)}$ then
$D \not\vdash \pd{\neg inf^{+}(r)}$
and even when $A(r)$ is proved defeasibly $p$, the consequent of $r$,
cannot be established defeasibly using $r$
because the link between the two is broken.

The transformation to eliminate defeaters is slightly similar 
to the above transformation;
two propositions are introduced for every proposition in $D$
and up to three rules are used to replace each rule in $D$.
The two propositions -- $p^+$ and $p^-$ -- introduced
for each proposition $p$,
represent that $p$ (respectively $\neg p$) is not defeated by a defeater.
Again, these propositions form links between
antecedent and consequent,
and the effect of a defeater is obtained with a defeasible rule that
can break such a link.
For more details, see \cite{ABGM99}.

It is clear that these two transformations are profligate
in their introduction of propositions and generation of rules.
Nevertheless, they can be applied in one pass (each) over
a defeasible theory, and the resulting defeasible theory
is, at most, a constant factor larger.

The following result was proved in \cite{ABGM99},
albeit in parts and with slightly different terminology.

\begin{thm}[\cite{ABGM99}]
Let $D$ be a defeasible theory and 
let $\Sigma$ be the language of $D$.
Let
$D' = \Basic{}(D)$.

\noindent
Then $D'$ is a basic defeasible theory and,
for all conclusions $c$ in $\Sigma$,

$D \vdash c$ iff $D' \vdash c$.

\noindent
Furthermore,
$D'$ can be constructed in time linear in the size of $D$, and
the size of $D'$ is linear in the size of $D$.
\end{thm}

We must also ensure that the basic defeasible theory 
resulting from \Basic{} has duplicated strict rules.
To guarantee this property we employ the following simple
transformation \linebreak
\DupStrict{}\footnote{
This duplication is not strictly necessary since an appropriate separation
of definite and defeasible reasoning is achieved by the transformations
of \cite{ABGM99}.  However, in this context, it is simpler to ensure
this property by duplication
than to argue that it holds of transformations that are not presented here.
}.

\begin{defn}
Let $D = (F, R, >)$.
Then $\DupStrict{}(D) = (F, R', >)$ where

\[
R' = R \cup 
\{ (r : B \Rightarrow q) ~|~ (r : B \rightarrow q) \in R_s \} \\
\]

\end{defn}

It is straightforward to show that \DupStrict{} does not change
the conclusions of a theory.

\begin{propn}
For any defeasible theory $D$, and any conclusion $c$,

$D \vdash c$ iff $\DupStrict{}(D) \vdash c$
\end{propn}
\begin{proof}
(Sketch)
Definite provability is affected only by strict rules and facts.
The strict rules in $R$ and $R'$ are essentially the same,
and the facts are unchanged.
Thus definite provability is unchanged by \DupStrict{}.

In applying the defeasible inference rules,
it makes no difference whether a rule is strict or defeasible.
So $\vdash'$ applied to $\DupStrict{}(D)$
uses effectively the same rules as
It also makes no difference whwther there is one or several copies of
a rule.
\end{proof}

\subsection{The Algorithm}

\newcommand{\while}{{\bf while\,\,\,}}
\newcommand{\whenever}{{\bf whenever\,\,\,}}
\newcommand{\case}{{\bf case\,\,\,}}
\newcommand{\tend}{{\bf end\,\,\,}}
\newcommand{\delete}{{\bf delete\,\,\,}}

\begin{figure}
\begin{tabbing}
1234\=1234\=1234\=1234\=1234\=1234\=\kill

$D'$ = $(F', R', \emptyset)$ = \Basic{}($D$) \\
$R$ = \DupStrict{}( $R'$ ) \\
initialize $S$ \\
$K$ = $\emptyset$ \\
 \\
\while ( $S \neq \emptyset$ ) \\
\>	choose $s \in S$ and delete $s$ from $S$ \\
\>	add $s$ to $K$ \\
\>	\case $s$ of \\
\>\>		$\PD{p}$: \\
\>\>\>			\delete all occurrences of $p$ in all rule bodies \\
\>\>\>			\whenever a body of a strict rule with head $h$ becomes empty \\
\>\>\>\>				add $\PD{h}$ to $S$ \\
\>\>\>\>				record $\PD{h}, \pd{h}, \ps{h}$ \\
\>\>\>\>				\CheckInference{}( $\PD{h}, S$ ) \\
\>\>\>			\whenever a body of a defeasible rule with head $h$ becomes empty \\
\>\>\>\>				record $\ps{h}$ \\
\>\>\>\>				\CheckInference{}( $\ps{h}, S$ ) \\
\>\>		$\MD{p}$: \\
\>\>\>			\delete all strict rules where $p$ occurs in the body \\
\>\>\>			\whenever there are no more strict rules for a literal $h$,
											and there is no fact $h$ \\
\>\>\>\>				add $\MD{h}$ to $S$ \\
\>\>\>\>				record $\MD{h}$ \\
\>\>\>\>				\CheckInference{}( $\MD{h}, S$ ) \\
\>\>		$\pd{p}$: \\
\>\>\>			\delete all occurrences of $p$ in defeasible rule bodies \\
\>\>\>			\whenever a body with head $h$ becomes empty \\
\>\>\>\>				record $\ps{h}$ \\
\>\>\>\>				\CheckInference{}( $\ps{h}, S$ ) \\
\>\>		$\md{p}$: \\
\>\>\>			\delete all defeasible rules where $p$ occurs in the body \\
\>\>\>			\whenever there are no more defeasible rules for a literal $h$ \\
\>\>\>\>				record $\ms{h}$ \\
\>\>\>\>				\CheckInference{}( $\ms{h}, S$ ) \\
\>	\tend \case \\
\tend \while \\
\end{tabbing}
\caption{Inference algorithm for defeasible logic}
\label{algm}
\end{figure}

In the algorithm, which is presented in Figure \ref{algm},
$p$ ranges over literals and
$s$ ranges over conclusions.
$K$ and $S$ are sets of conclusions.
$K$ accumulates the set of conclusions that have been proved and used,
while $S$ holds those proven conclusions that have not yet
been used to establish more conclusions.
$D$ is the input defeasible theory.

To begin the algorithm we 
reduce $D$ to basic defeasible logic with duplicated strict rules,
as discussed above.
We also initialize the set $S$, that is, we add to $S$
those conclusions that can immediately be established:
all facts are definitely true,
as are the heads of strict rules with empty bodies.
The heads of defeasible rules with empty bodies are tentatively true.
Similarly, those literals with no (strict) rules for them
are unprovable (strictly).

The algorithm proceeds by modifying the rules in the theory.
For strict rules, when inferring positive definite consequences,
the algorithm is similar to unit resolution for definite clauses
in classical logic:
when an atom is proved, it can be eliminated from the bodies of
all other definite clauses.
In this case, when a literal is established definitely
it can be deleted from the body of all rules.
Similarly, when it is established that a literal $p$ cannot be proved
then those strict rules which have $p$ as a pre-condition cannot
be used to prove the head,
and so they can be deleted.
When a strict rule has an empty body, then its head is definitely provable;
when there are no strict rules for a literal, then the literal is
unprovable definitely.
When an atom is proved definitely it is eliminated from both
strict and defeasible rules;
this necessitates extra code for the case when the body of a defeasible
rule becomes empty.

When establishing tentative provability,
we proceed in exactly the same way as with definite provability,
except that we restrict attention to defeasible rules.
We can conclude $\ps{q}$ precisely when the body of a defeasible rule for $q$
becomes empty,
and $\ms{q}$ precisely when there are no more defeasible rules for $q$.

When establishing defeasible provability,
we use the simplified inference rules introduced at the end of
Section \ref{sec:dl}.
Each time a statement such as $\ps{p}$ is inferred by the system
the statement is {\em record}ed in the data structure for $p$
and we check to see whether either
of the simplified inference rules for $\partial$ can be applied.
This task is performed by \CheckInference{},
which will add
either $\pd{p}$ or $\md{p}$, if justified, to the set $S$\footnote{
  Recall that defeasible logic will never infer both 
  $\pd{p}$ and $\md{p}$ since it is coherent \cite{Billington93}.
}.
For example, \CheckInference{} ($\ps{p}, S$)
will check whether $\MD{\non p}$ and $\ms{\non p}$
have already been established,
so that $\pd{p}$ may be inferred and added to $S$,
and if $\MD{\non p}$ has been established will add $\md{\non p}$ to $S$.
Similarly,
\CheckInference{}($\MD{p}, S$)
will check whether $\ps{\non p}$ and $\ms{p}$,
so that $\pd{\non p}$ might be inferred,
and will check $\ms{p}$, $\PD{\non p}$ and $\ps{\non p}$
to decide whether $\md{p}$ might be proved.

Execution of this algorithm can be understood as execution of
the transition system of Section \ref{trans}.
$S \cup K$ combined with the set of recorded extended conclusions
in the algorithm corresponds to $C_i$ in the transition system.
The deletions in the algorithm correspond to transitions
\ref{delPD}, \ref{delpd}, \ref{delMD}, and \ref{delmd}.
The add statements correspond to transitions
\ref{PD} and \ref{MD}.
The record statements correspond to transitions
\ref{PD}, \ref{ps}, \ref{MD}, \ref{ps}, and \ref{ms}.
These transitions are also reflected in the initialization of $S$.
Finally,
\CheckInference{} embodies transitions \ref{inferpd} and \ref{infermd}.

The algorithm corresponds to a restricted form of the transition system
where some sequences of transitions are disallowed.
For example,
the first case of Figure \ref{algm}
requires that all transitions \ref{delPD} and \ref{delpd} involving $p$
must occur as a block,
uninterrupted by any other uses of these transitions.
Furthermore, the only possible other transitions in the block
are transitions \ref{PD} and \ref{ps},
when $p$ is the last remaining literal in a rule,
and transitions \ref{inferpd} and \ref{infermd}
when the recorded information about $p$ triggers them.
Note, however, that the algorithm does not disallow any transitions;
it simply restricts the order in which transitions may occur.
Thus the soundness and completeness of the transition system
extends to soundness and completeness of the algorithm.

\begin{figure}[ht]
\psfig{figure=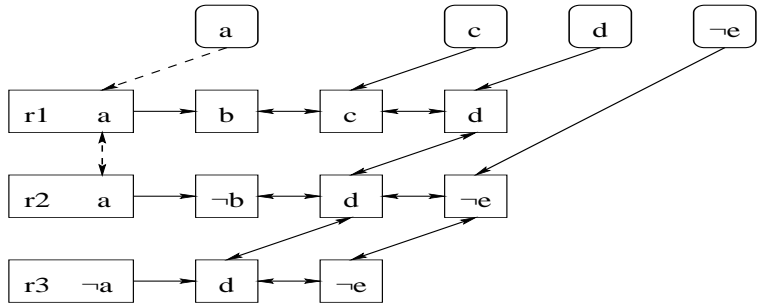,width=10cm,height=4cm}
\caption{Data Structure for Rules}
\label{dspic}
\end{figure}

The key to an efficient implementation of this algorithm
is the data structure used to represent the rules.
It is exemplified (albeit incompletely) in
Figure \ref{dspic}
for the theory
\[
\begin{array}{lrr}
r_1: & b, c, d \Rightarrow & a \\
r_2: & \neg b, d, \neg e  \Rightarrow & a \\
r_3: & d, \neg e  \Rightarrow & \neg a \\
\end{array}
\]

Each rule body is represented as a doubly-linked list
(horizontal arrows in Figure \ref{dspic}).
Furthermore, for each literal $p$ there are doubly-linked lists
of the occurrences of $p$ in the bodies of strict
(respectively, defeasible) rules (diagonal arrows).
For each literal $p$,
there are doubly-linked lists of the strict
(respectively, defeasible) rules for $p$
(dashed arrows).
Each literal occurrence has a link to the record for the rule it
occurs in
(not shown in Figure \ref{dspic}).
As mentioned above, each literal $p$ holds information
showing which extended conclusions about $p$ have been established,
and there is also a link to the corresponding data structure for $\:\non p$.

\input{psfig}

This data structure allows the deletion of literals and rules
in time proportional to the number of literals deleted.
Furthermore, we can detect in constant time whether a literal
deleted was the only literal in that body,
and whether a rule deleted with head $h$
was the only strict (defeasible) rule for $h$.
Each literal occurrence is deleted at most once,
and the test for empty body is made at most once per deletion.
Similarly, each rule is deleted at most once,
and the test for no more rules is made once per deletion.
The cost of a call to \CheckInference{} is constant,
since all checking relies on data that is immediately available,
and at most 3 checks are needed per call.
Thus the cost of the main part of the algorithm is $O(N)$,
where $N$ is the number of literal occurrences in $D'$.

Furthermore, the initialization of $S$ is supported by the data structure
and has cost proportional to the number of propositions in $D'$.
Note that $S$ can be implemented as a queue or a stack,
since the only operations on $S$ are
to test for emptiness,
add an element, or
choose and delete an element.
These operations require only constant time,
and the number of operations is, in the worst case,
proportional to the number of rules in $D'$.
Since the initial transformations \cite{ABGM99} that produce $D'$
are linear, the cost of the entire algorithm is linear in the size of $D$.

\begin{thm}
The consequences of a defeasible theory $D$
can be computed in $O(N)$ time,
where $N$ is the number of symbols in $D$.
\end{thm}

This algorithm, when restricted to positive definite conclusions,
is similar to the bottom-up linear algorithm
for determining satisfiability of Horn clauses of
Dowling and Gallier \cite{Dowling84,GU89}.
One difference is in the data structures:
the Dowling-Gallier algorithm keeps a count of the number of atoms in
the body of a rule, rather than keep track of the body.
The latter results in greater memory usage,
but allows us to reconstruct the residue of the computation:
the simplified rules that remain.
This residue is useful in understanding the behaviour of a theory,
and can be useful in some applications of defeasible logic.

\section{Discussion} \label{disc}

There seem to be several inter-related features of defeasible logic
that contribute to its linear complexity:

\begin{itemize}
\item form of failure \\
The form of failure-to-prove exhibited in the proof conditions
for $\MD$ and $\md$ is closely related to Kunen's semantics
for logic programs with negation-as-failure \cite{MG99}.
This form of failure does not include looping failures,
and the propositional form is known to have linear complexity.

\item static and local nature of conflict \\
When trying to derive $\pd{q}$ or $\md{q}$ it is sufficient
to look at rules for $q$ and $\non q$,
since $\non q$ is the only literal that conflicts with $q$.
Thus the set of conflicting rules is unchanging and readily apparent.
In logics where there is no bound on the number of conflicting literals
and/or the conflicting rules cannot be known {\em a priori}
(for example, plausible logic \cite{plausible})
the corresponding inference rules for $\pd$ and $\md$ can be expected to be
more complex.

\item linear cost elimination of superiority relation \\
Defeaters can be handled as a variant of defeasible rules
using the techniques of Section \ref{acdl}.
The presence of a superiority relation complicates
these techniques greatly,
although it is clear that there is a linear direct implementation
of full defeasible logic by imitating the elimination transformation.
The elimination of the superiority relation is based on
a simple syntactic transformation \cite{ABGM99},
but a similar transformation has been proposed for
other logics \cite{KT}
that have higher complexity.
The success of such a simple transformation can be partly attributed
to the static nature of the superiority relation
and the static and local nature of conflicting literals,
which do not change over the course of evaluation.
\end{itemize}

A significant non-factor in the complexity of defeasible logic
is the team defeat aspect of the logic.
This might be expected to create some difficulties since,
conceptually, two teams of rules are pitted against each other,
and this might give rise to combinatorial problems.
However, reformulating the issue as one of competing/conflicting
literals clarifies that there is a fixed number of pieces of
information that are of interest, and each rule can contribute
individually.

We can expect that similar logics,
such as the variants discussed in \cite{ABGM00}
and variants where strict rules are superior to defeasible rules \cite{Nute94},
also have linear complexity and are amenable to the techniques used here,
although the details will require careful verification.
Similarly, well-founded defeasible logic \cite{MG99} can be expected
to have quadratic complexity,
since it employs the well-founded semantics \cite{VGRS} notion of failure,
which has quadratic complexity.

\section{Conclusion} \label{conc}

We have shown that propositional defeasible logic has linear complexity.
The algorithm that we have presented here has been implemented 
as the system Delores \cite{Miller}
and has undergone a preliminary experimental evaluation \cite{MRABM}.
It executes simple, but large, theories of basic defeasible logic
very quickly.
The linear complexity of the algorithm is supported empirically.

However, the transformations used to convert an arbitrary defeasible
theory to the appropriate form for the algorithm of Figure \ref{algm}
impose a large constant factor
on the cost of initializing $S$.
Although the cause has not been pinpointed,
it appears to be derived from the multiplication of rules and propositions
during the transformations.
It should be possible to improve performance by integrating the
transformations of \cite{ABGM99} more tightly and/or
abandoning the design goal \cite{ABGM99} of
incrementality of the transformations
(with respect to changes in the original defeasible theory).

Work is proceeding on improvements to the implementation of defeasible logic
and the application of similar techniques to implement
well-founded defeasible logic.
We propose to use the implementation to support reasoning about
regulations \cite{ABM99}.

\section*{Acknowledgements}
Thanks to Grigoris Antoniou, David Billington,
Guido Governatori and Andrew Rock
for discussions on defeasible logic and its implementation.
Thanks to Tristan Miller for discussion of the algorithm and
his implementation of Delores.
Also thanks to the anonymous referees for their careful reading
and sugestions.
This research was supported by the
Australian Research Council.

\end{document}